\documentclass{mva_style}
\usepackage{graphicx}
\usepackage{diagbox}
\usepackage[pagebackref,breaklinks,colorlinks,bookmarks=false]{hyperref}
\usepackage{amsmath}
\usepackage{amssymb}
\usepackage[labelformat=simple]{subcaption}
\usepackage{multirow}
\usepackage{multicol}
\usepackage{diagbox}
\usepackage{cite}

\newcommand{\R}{\mathbb{R}}

\finalcopy

\begin{document}
\title{Joint learning of images and videos\\ with a single Vision Transformer}

\author{
Shuki Shimizu\\
Nagoya Institute of Technology\\
Nagoya, Japan\\
{\tt s.shimizu.562@nitech.jp}
\and
Toru Tamaki\\
Nagoya Institute of Technology\\
Nagoya, Japan\\
{\tt tamaki.toru@nitech.ac.jp}\\
}

\maketitle

\section*{\centering Abstract}
\textit{

In this study, we propose a method for jointly learning of images and videos using a single model.
In general, images and videos are often trained by separate models.
We propose in this paper a method that
takes a batch of images as input to Vision Transformer 
(\emph{IV-ViT}), and also a set of video frames with temporal aggregation by late fusion.
Experimental results on two image datasets and two action recognition datasets are presented.
}

\section{Introduction}

Image and action recognition are long-standing tasks and have been studied extensively. 
Many deep learning models for image recognition have been proposed\cite{Dosovitskiy_ICLR2021_ViT_Vision_transformer, He_CVPR2016_ResNet, Bulat_NeurIPS2021_X-ViT,Simonyan_ICLR2015_VGG,Szegedy_CVPR2015_GoogLeNet_Inception}, and many datasets for image recognition have been made publicly available\cite{Deng_CVPR2009_ImageNet, LeCun_IEEE1998_MNIST_LeNet}.
Action recognition also has a long history\cite{Hara_IEICE_ED2020_Action_Recognition_Survey, Yu_2022_human_action_recog_survey ,Zhu_arXiv2020_Action_Recognition_Survey},
with large datasets\cite{Kuehne_ICCV2011_HMDB51, Li_arXiv2020_AVA-Kinetics,Soomro_arXiv2012_UCF101}
and various models\cite{Arnab_2021_ICCV_ViVit, Carreira_2017CVPR_I3D,Feichtenhofer_2020CVPR_X3D,Feichtenhofer_2019ICCV_SlowFast,Hara_2018CVPR_3D_ResNet,Simonyan_NIPS2014_Two-Stream,Tran_2015ICCV_C3D,Zhang_ACMMM2021_TokenShift} proposed.

With the rapid growth of Vision Transformer (ViT)\cite{Dosovitskiy_ICLR2021_ViT_Vision_transformer} in recent years, research on learning a multi-modal model for images and videos has emerged\cite{Fan_ICLR2022_SIFAR,Girdhar_CVPR2022_Omnivore,Likhosherstov_arXiv2021_PolyViT,Girdhar_arXiv2022_OmniMAE}. 
Although many CNN-based image and action recognition models have been proposed, most of them were developed separately because of the modality difference between images and videos in terms of temporal modeling. 
As ViT became popular, action recognition models based on ViT were proposed \cite{Selva_ArXiv2022_Video_Trans_Survey}. 
ViT divides an image into patches and applies Transformer \cite{Vaswani_arXiv2017_transformer} by considering the patches as tokens, while ViT-based action recognition models divide a video into patches.
This structure allows models to simplify the model design by providing both images and videos in a similar way \cite{Likhosherstov_arXiv2021_PolyViT,Girdhar_CVPR2022_Omnivore}.

\begin{figure}
    \centering
    
    \hfil
    \begin{minipage}[m]{0.3\linewidth}
        \centering
        \includegraphics[width=\linewidth]{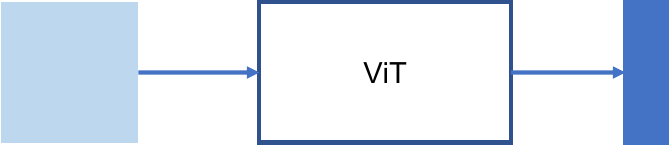}
        \subcaption{}
    \end{minipage}
    \hfil
    \begin{minipage}[m]{0.5\linewidth}
        \centering
        \includegraphics[width=\linewidth]{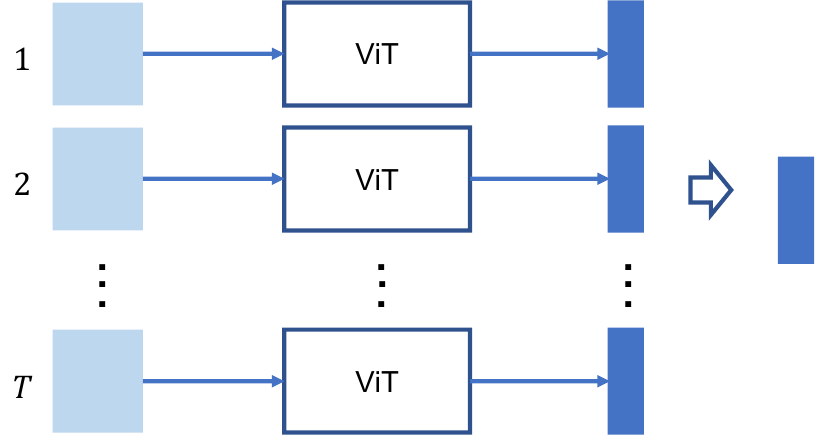}
        \subcaption{}
    \end{minipage}
    \hfill
    
    \caption{The proposed IV-ViT for cases
    (a) when input is an image, and 
    (b) when input is a video clip of $T$ frames.
    }
    \label{fig:proposed model overview}
\end{figure}

However, prior works have not fully exploited the relation of image and video modalities;
an image is considered a video with one frame, while a video is a series of frames.
In this paper, we propose \emph{IV-ViT}, a model for joint learning of images and videos
with a single ViT model with a maximal parameter sharing.
The proposed method shown in Fig.~\ref{fig:proposed model overview} 
applies a ViT to each frame, which is a frame-wise ViT often used as a baseline of action recognition\cite{Zhang_ACMMM2021_TokenShift,Hashiguchi_2022_ACCVW_MSCA}.
When recognizing a batch of images, ViT is simply applied, and when recognizing a video, the output of ViT for each frame of the video is aggregated by late fusion\cite{Karpathy_2014_CVPR_Sports-1M}.
This structure enables joint learning of images and videos using a single ViT model.
To enable the frame-wise model to handle the temporal information, we utilize the temporal feature shift
\cite{Hashiguchi_2022_ACCVW_MSCA,Lin_2019ICCV_TSM,Zhang_ACMMM2021_TokenShift}
by interacting features between neighbor frames.

\section{Related work}

\subsection{Image and action recognition}

Since ImageNet \cite{Deng_CVPR2009_ImageNet},
many CNN models \cite{He_CVPR2016_ResNet,Krizhevsky_NeurIPS2012_AlexNet,Simonyan_ICLR2015_VGG,Szegedy_CVPR2015_GoogLeNet_Inception}
and recently ViT models
\cite{Dosovitskiy_ICLR2021_ViT_Vision_transformer,Liu_2021_ICCV_Swin_Transformer}
have been proposed for image recognition.
There has also been a lot of research on action recognition
based on 3D CNN
\cite{Carreira_2017CVPR_I3D,Feichtenhofer_2020CVPR_X3D,Feichtenhofer_2019ICCV_SlowFast,Hara_2018CVPR_3D_ResNet,Tran_2015ICCV_C3D}
and ViT
\cite{Arnab_2021_ICCV_ViVit,Neimark_2021_ICCVW_VideoTransformerNetwork,Sharir_arxiv2021_STAM,Zhang_ACMMM2021_TokenShift}.

For efficient modeling of temporal information with 3D CNN,
some works used spatio-temporal separable convolutions instead of fully 3D convolutions
\cite{Tran_2018CVPR_R(2+1)D,Xie_2018ECCV_S3D_Kinetics200,Qiu_2017ICCV_P3D},
and others introduced temporal feature shift in frame-wise (or 2D CNN-based) models
\cite{Lin_2019ICCV_TSM,Sudhakaran_CVPR2020_GatedShiftNetwork,Fan_ECCV2020_RubiksNet}.
ViT-based video models have the self-attention mechanism to model the temporal information, such as using transformer for late-fusion \cite{Neimark_2021_ICCVW_VideoTransformerNetwork,Sharir_arxiv2021_STAM}, separate spatial and temporal attentions \cite{Arnab_2021_ICCV_ViVit}, or restricted spatio-temporal attention \cite{Bertasius_ICML2021_TimeSformer,Bulat_NeurIPS2021_X-ViT}, and feature shift \cite{Zhang_ACMMM2021_TokenShift,Hashiguchi_2022_ACCVW_MSCA}.

\subsection{Joint learning of images and videos}

Several attempts have been made to recognize both images and videos with a single model even using CNN.
UniDual\cite{Wang_arXiv2019_UniDual} and Omnisorce\cite{Duan_arXiv2020_OmniSource} 
used 3D CNN models \cite{Feichtenhofer_2019ICCV_SlowFast,Tran_2018CVPR_R(2+1)D} for both videos and images
by first replicating a still image $T$ times
and then converting them to a 3D volume of $T$ frames,
which is inefficient for learning images.

Unlike CNN, ViT uses patch embeddings, which makes it easier to handle 
different modalities in a unified manner. 
PolyViT \cite{Likhosherstov_arXiv2021_PolyViT} and Omnivore \cite{Girdhar_CVPR2022_Omnivore} simultaneously learn multi-modal data including images, videos, and audio \cite{Likhosherstov_arXiv2021_PolyViT} or RGB+Depth \cite{Girdhar_CVPR2022_Omnivore}.

Omnivore\cite{Girdhar_CVPR2022_Omnivore} uses the same embedding layer
for both images and videos, however,
by zero-padding to a still image $T$ times
and then converting them to a 3D volume of $T$ frames,
which is as inefficient as \cite{Wang_arXiv2019_UniDual,Duan_arXiv2020_OmniSource}.
PolyViT \cite{Likhosherstov_arXiv2021_PolyViT} uses 
different embedding layers for different modalities,
even for images and videos, although both are the same visual modality.

In contrast, our method, IV-ViT, shares the same embedding layer
for images and videos like as 
Omnivore\cite{Girdhar_CVPR2022_Omnivore},
but without converting images into 3D volumes by replication or zero-padding.
Instead, we train a single model for a batch of images
and a batch of video frames in the same manner, as explained below.

\section{ViT model}

In this section, we give a brief introduction to ViT for images \cite{Dosovitskiy_ICLR2021_ViT_Vision_transformer} and a frame-wise ViT for videos \cite{Zhang_ACMMM2021_TokenShift}.

\subsection{Image}

Let $x \in \R^{3 \times H \times W}$ be an input image, where $H$ and $W$ are the height and width of the image. The input image is first divided into $P \times P$ pixel patches and transformed into a tensor $\hat{x} = [x_0^1, \ldots, x_0^N] \in \R^{N \times d}$, where $x_0^i \in \R^{d}$ is the $i$-th patch, $N = \frac{HW}{P^2}$ is the number of patches, and $d = 3 P^2$ is the patch dimension. The input patch $x_0^i$ is transformed by a linear transformation using the $D$-dimensional embedding matrix $E$ and the addition of the positional embedding $E_\mathrm{pos}$ as follows;
\begin{align}
    z_0 &= [c_0, x_0^1 E, x_0^2 E, \ldots, x_0^N E] + E_\mathrm{pos},
\end{align}
where $c_0 \in \R^{D}$ is a class token. This patch embedding $z_0 \in \R^{(N+1) \times D}$ is input to the first encoder block $M^\ell$ of the transformer, in which the self-attention mechanism is applied to patches.
Let $z_\ell$ be the input to the $\ell$-th block $M^\ell$ and $z_\ell$ be the output;

$
    z_{\ell} = M^\ell (z_{\ell - 1})
$.
Then, the output $z_L$ of ViT consisting of $L$ blocks is
\begin{align}
    z_L = (M^L
    \circ \cdots M^1 \circ M^0) (x)
    \in \R^{(N+1) \times D}.
\end{align}
Note that $\circ$ is the composition and $M^0$ is the embedding layer. The final head $M^h(z_L)$

often uses the part of $z_L$ corresponding to the class token only;

$
    \tilde{z}_{L, d} = z_{L, 0, d}
$,
where $(n,d)$ elements of $z_\ell$ is denoted by $z_{\ell, n, d}$.
Then the class prediction is
\begin{align}
    \hat{y} &= M^h(\tilde{z}_L) \in [0, 1]^C,
\end{align}
where $C$ is the number of categories.

\subsection{Video}

Let $x \in \R^{T \times 3 \times H \times W}$ be an input video clip where $T$ is the number of frames in the video clip. Each frame is transformed separately into
patches like images.

The differences are the dimension of the class token $c_0 \in \R^{T \times D}$ and patch embeddings $z_0 \in \R^{T \times (N+1) \times D}$.
The output $z_L$ of the same ViT with $L$ blocks $M^{\ell} (\ell=0,\ldots,L)$ is as follows.
\begin{align}
    z_L = (M^L
    \cdots M^1 \circ M^0) (x)
    \in \R^{T \times (N+1) \times D}.
\end{align}

We use a simple average pooling
\cite{Hashiguchi_2022_ACCVW_MSCA,Zhang_ACMMM2021_TokenShift}
for temporal aggregation to summarize outputs from each frame of the video.
Let $(t,n,d)$ elements of $z_\ell$ be denoted by $z_{\ell,t, n, d}$ and $\tilde{z}_\ell \in \R^{T \times D}$ corresponding to the class token be denoted by $\tilde{z}_{\ell, t, d}$. The class token
$
    \tilde{z}_{L, t, d} = z_{L, t, 0, d}
$
is aggregated by
\begin{align}
    \bar{z}_{L, d} &= \frac{1}{T} \sum_t^T \tilde{z}_{L, t, d},
\end{align}
and the head $M^h$ predicts the category by
\begin{align}
    \hat{y} &= M^h(\bar{z}_L) \in [0, 1]^C.
\end{align}

\section{Proposed IV-ViT Model}

\subsection{Model input}

The above ViT for images and videos can be handled in a consistent manner. In general when training with images, a batch of images $x \in \R^{B \times 3 \times H \times W}$ of size $B$ is used as input, and the output $\hat{y} \in [0, 1]^{B \times C}$ is for the batch. In other words, the shape of input/output of ViT blocks $M_\ell$ is the same for images and videos,
and the ViT blocks do not need to distinguish whether the input is video frames or a batch of images.
That is, the output of the $\ell$-th block is $z_\ell \in \R^{B \times (N+1) \times D}$ for a batch of images,
and $z_\ell \in \mathbb{R}^{BT \times (N+1) \times D}$ for a batch of video frames.
Therefore, the same ViT architecture can be used for images and videos, including the embedding layer.

\noindent\textbf{Images: }
Take a batch of images $x \in \mathbb{R}^{B \times 3 \times H \times W}$ as input. 
The output of the $L$-th block is $z_L \in \R^{B \times (N+1) \times D}$, 
and the class token $\tilde{z}_L \in \R^{B \times D}$ is used to predict $\hat{y} \in [0, 1]^{B \times C}$ by the head.
This is a common case when training a ViT for images.

\noindent\textbf{Videos: }
Take a batch of video clips $x \in \mathbb{R}^{B \times T \times 3 \times H \times W}$ as input.
To input this 5D tensor to ViT, the time dimension is expanded to the batch dimension and transformed into a 4D tensor $x' \in \mathbb{R}^{BT \times 3 \times H \times W}$.
The output of the $L$-th block $z_L \in \mathbb{R}^{BT \times (N+1) \times D}$ is first transformed into $z'_L \in \mathbb{R}^{B \times T \times (N+1) \times D}$. Then, the class token part $\tilde{z}'_L \in \mathbb{R}^{B \times T \times D}$ is extracted and average pooling over the time dimension is applied to obtain $\bar{z}'_L \in \mathbb{R}^{B \times D}$. The output by the head is $\hat{y} \in [0, 1]^{B \times C}$.

\subsection{Modeling of temporal information}

\begin{figure}[t]

    \centering
    \includegraphics[width=0.7\linewidth]{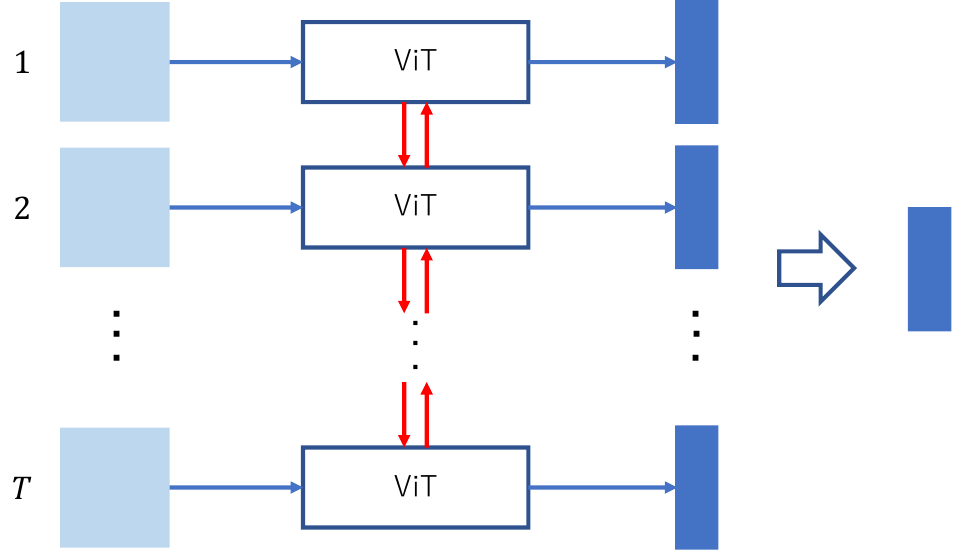}
    \caption{
    Concept of feature shift between ViT modules in successive frames.
    }
    \label{Fig:feature_shift}

\end{figure}

The proposed IV-ViT model uses a frame-wise ViT for videos,
which is not sufficient to model the temporal information of the video.
Therefore, we introduce the feature shift \cite{Zhang_ACMMM2021_TokenShift,Hashiguchi_2022_ACCVW_MSCA,Bulat_NeurIPS2021_X-ViT} that can handle temporal information while maintaining the structure, as shown in Fig.~\ref{Fig:feature_shift}.

For the frame dimension $t$ of features $z_{t, n, d}^{\ell}$ in the $\ell$-th block $M^\ell$, we shift elements between successive frames $t+1$ and $t-1$ by a certain amount $D_b$ and $D_f$ as follows.

\begin{align}
    z_{\mathrm{out}, t, 0, d}^{\ell}
        &= 
    \begin{cases}
        z_{\mathrm{in}, t-1, 0, d}^{\ell}, & \text{$1 < t \le T, 1 \le d < D_b$}\\
        z_{\mathrm{in}, t+1, 0, d}^{\ell}, & \text{$1 \le t < T, D_b \le d < D_b+D_f$}\\
        z_{\mathrm{in}, t, 0, d}^{\ell},   & \text{$\forall t, D_b+D_f \le d \le D$}\\
    \end{cases}\\
    z_{\mathrm{out}, t, n, d}^{\ell}
        &= z_{\mathrm{in}, t, n, d}^{\ell}, \hspace{2.7em} \text{$\forall t, 1 \le n < N, \forall d$},
\end{align}

where $z_{\mathrm{in}, t, n, d}^{\ell}$ and $z_{\mathrm{out}, t, n, d}^{\ell}$ are features $z_{t, n, d}^{\ell}$ before and after the shift operation,
which is the same as TokenShift\cite{Zhang_ACMMM2021_TokenShift}.

In this way, one block shifts the information in the time direction for one time period. Therefore, it is possible to interact the temporal information of the whole clip by using the number of ViT blocks $L$ larger than the clip length $T$ \cite{Hashiguchi_2022_ACCVW_MSCA}.

\begin{table*}[t]
    \centering
    \caption{Performance comparison for four image and video datasets. Note that averages for ``IV-ViT (images/videos)'' are for each domain.}
    \label{Tab:comparison_with_polyvit} 
    \begin{tabular}{cc|cccccccc|cc}
        \multicolumn{2}{c}{} & \multicolumn{2}{c}{Tiny-ImageNet} & \multicolumn{2}{c}{CIFAR100} &  \multicolumn{2}{c}{UCF101} &  \multicolumn{2}{c}{mini-Kinetics} &  \multicolumn{2}{c}{average} \\
             & pretrain & top-1 & top-5 & top-1 & top-5 & top-1 & top-5 & top-1 & top-5 & top-1 & top-5 \\ 
        \hline
        IV-ViT & K400 & 83.22 & 94.97 & 88.69 & 98.61 & 90.08 & 98.81 & 62.83 & 85.55 & 81.21 & 94.49\\
        IV-ViT (separate) & K400 & 83.10 & 95.18 & 88.89 & 98.71 & 89.34 & 98.89 & 62.71 & 85.03 & 81.01 & 94.45 \\
        IV-ViT (images) & K400 & 83.95 & 95.44 & 87.74 & 98.49 & -- & -- & -- & -- & \textcolor[gray]{0.5}{85.84} & \textcolor[gray]{0.5}{96.96}\\
        IV-ViT (videos) & K400 & -- & -- & -- & -- & 93.10 & 99.39 & 68.75 & 88.42 & \textcolor[gray]{0.5}{80.92} & \textcolor[gray]{0.5}{93.90} \\
        PolyViT & K400 & 80.82 & 94.18 & 86.33 & 98.20 & 92.23 & 99.02 & 71.49 & 89.98 & 82.72 & 95.34\\
        PolyViT & K400+I21K & 80.29 & 93.84 & 86.36 & 97.88 & 91.88 & 98.84 & 71.68 & 89.74 & 82.55 & 95.07\\ \hline
        IV-ViT & scratch & 23.65 & 48.57 & 35.23 & 66.54 & 41.06 & 73.10 & 13.27 & 34.31 & 28.30 & 55.63\\
        PolyViT & scratch & 14.88 & 35.60 & 24.30 & 53.57 & 32.09 & 59.64 & 10.54 & 27.55 & 20.45 & 44.09 \\
    \end{tabular}
\end{table*}

\section{Experimental results}

\subsection{Datasets}

In our experiments, we used the following two image and two video datasets, in total four datasets.

\begin{itemize}

\item Tiny-ImageNet \cite{Tiny_ImageNet_kaggle2018}:
A subset of ImageNet\cite{Deng_CVPR2009_ImageNet} consisting of 100,000 training images and 10,000 validation images. Each training image is $64 \times 64$ selected from 200 ImageNet classes (500 images per class).

\item CIFAR-100 \cite{Krizhevsky_TR2009_CIFAR10-100}:
An image dataset consisting of 50,000 training images and 10,000 validation images. The number of categories is 100 and the size is $32 \times 32$.

\item Mini-Kinetics (aka Kinetics200) \cite{Xie_2018ECCV_S3D_Kinetics200}:
A subset of Kinetics400 \cite{kay_arXiv2017_kinetics400},
an action recognition dataset, consisting of 78246 training videos and 5000 validation videos. Each video is selected from 200 classes of Kinetics400 (about 400 videos for each class).

\item UCF101 \cite{Soomro_arXiv2012_UCF101}:
A 101-class action recognition dataset consisting of 9537 training videos and 3783 validation videos.

\end{itemize}

\subsection{Model}

In experiments, we used ViT \cite{Dosovitskiy_ICLR2021_ViT_Vision_transformer}
pretrained on Kinetics400 \cite{kay_arXiv2017_kinetics400}
as a frame-wise ViT given by \cite{Zhang_ACMMM2021_TokenShift}.
We followed 
TokenShift\cite{Zhang_ACMMM2021_TokenShift} 
for temporal feature shift.
Hence the IV-ViT model has four heads for each dataset,
while the same embedding layer is used for all datasets.

\subsection{Training samples}

For images, 
a crop was made with a random area 
in the range $[0.08, 1.0]$ with respect to the area of the original image,
and a random aspect ratio in the range $[\frac{3}{4}, \frac{4}{3}]$,
and resize it to $224 \times 224$ pixels.
Then the crop was flipped horizontally with a probability of $1/2$.

For videos, a clip was created by extracting consecutive frames corresponding to a specified duration (2.67 seconds in experiments) from a video, and sampling $T=16$ frames uniformly from the extracted frames.
The shorter edge of the extracted frame was randomly resized in the range of $[224, 320]$ pixels  while maintaining the aspect ratio. Then $224 \times 224$ pixels at random locations were cropped and flipped horizontally with a probability of $1/2$.

Mini-batches were sampled from each dataset sequentially in a fixed order 
(Tiny-ImageNet, CIFAR100, UCF101, and mini-Kinetics),
with the batch size of 6 for both image and video datasets.
This means that a mini-batch of videos consists of 96 frames, since each of 6 video clips consists of 16 frames.
When a mini-batch was sampled from each dataset,
the gradient were computed and the parameters were updated.
The number of iterations was set to 20k, each iteration includes updates by four datasets
(so in total 80k iterations),
and epochs for each dataset do not align as in \cite{Omi_IEICE-ED2022_MDL}.
The optimizer was AdamW \cite{Loshchilov_ICLR2019_AdamW}
with the learning rate of 1e-5 and weight decay of 5e-5
for all datasets, although each dataset might have different optimal settings \cite{Likhosherstov_arXiv2021_PolyViT}.

\begin{figure}[t]

    \centering

    \begin{minipage}[t]{0.45\linewidth}
        \centering
        \includegraphics[width=\linewidth]{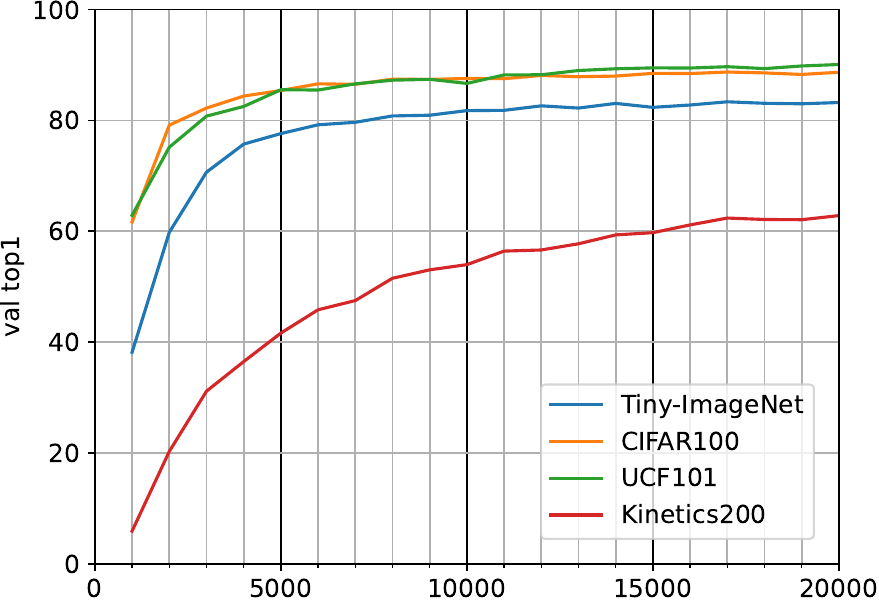}
        \subcaption{IV-ViT (K400)}
        \label{fig:ours}
    \end{minipage}
    \hfill
    \begin{minipage}[t]{0.45\linewidth}
        \centering
        \includegraphics[width=\linewidth]{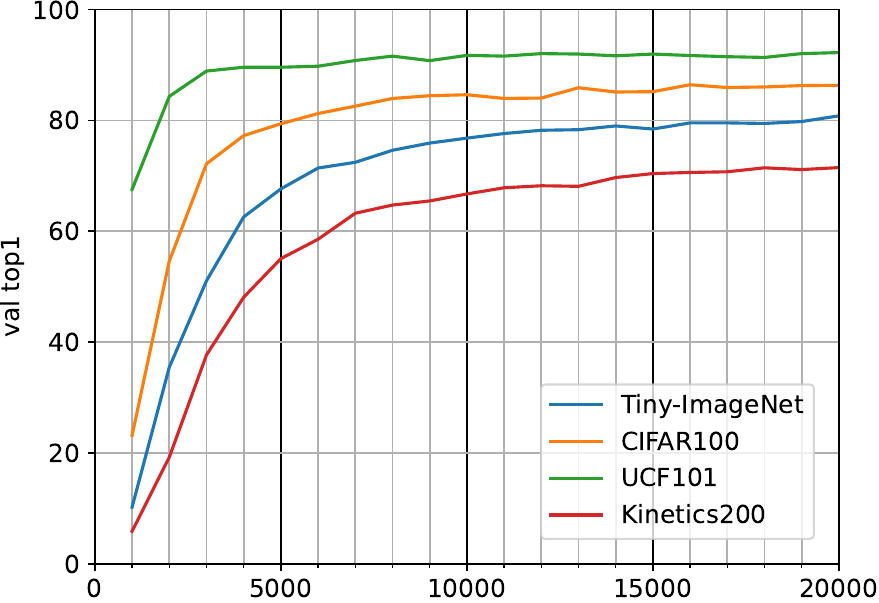}
        \subcaption{PolyViT (K400)}
        \label{fig:PolyViT K400}
    \end{minipage}
    
    \vspace{1em}
    
    \begin{minipage}[t]{0.45\linewidth}
        \centering
        \includegraphics[width=\linewidth]{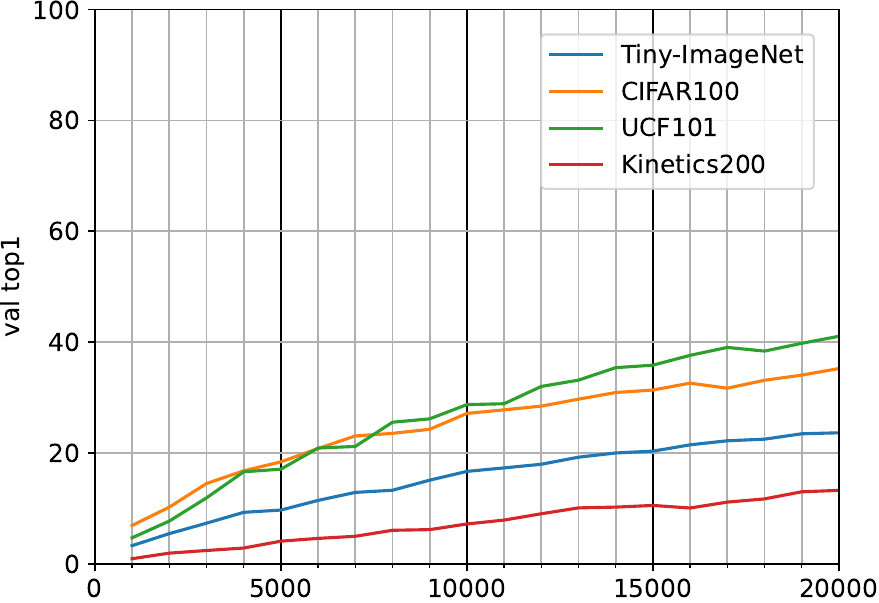}
        \subcaption{IV-ViT (scratch)}
        \label{fig:ours scratch}
    \end{minipage}
    \hfill
    \begin{minipage}[t]{0.45\linewidth}
        \centering
        \includegraphics[width=\linewidth]{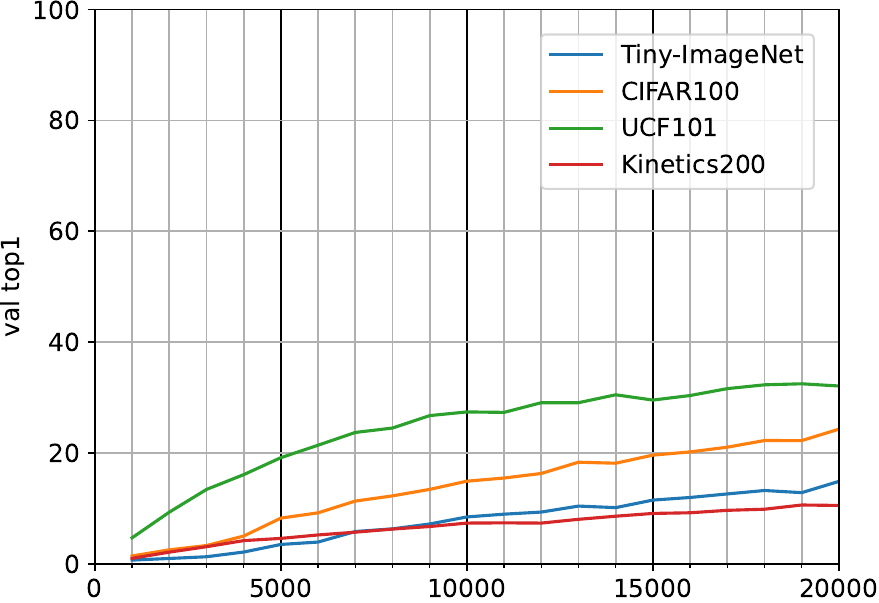}
        \subcaption{PolyViT (scratch)}
        \label{fig:PolyViT scratch}
    \end{minipage}

    \caption{
    Training progress of our model and PolyViT with and without pretraining.
    }
    \label{fig:compare_polyvit_pretrain_scratch}
\end{figure}

\subsection{Results and comparison}

Table~\ref{Tab:comparison_with_polyvit} shows the performance comparison.
The avg column shows the average of top-1/5 accuracy of four datasets.
We compared the proposed IV-ViT to PolyViT \cite{Likhosherstov_arXiv2021_PolyViT}
with our implementation.
For PolyViT, we used the same Kinetics400-pretrained weights with IV-ViT for the image embedding layer and the transformer blocks, while we used
Kinetics400-pretrained ViViT \cite{Arnab_2021_ICCV_ViVit} for the video embedding layer.
As an alternative, we added a version using ImageNet21k-pretrained weights for the image embedding layer, while others maintained.
The training procedure was the same with IV-ViT, except 
the number of frames in a clip was 32 and the batch size was set to 3.

The proposed IV-ViT performs better than PolyViT for images,
even when PolyViT uses ImageNet21k-pretrained weights for the image embedding layer.
Our model, however, works worse for videos, in particular mini-Kinetics.
This may be due to the use of the same embedding layer for images and videos in our model, while PolyViT uses different embedding layers.
However, as shown in the row ``IV-ViT (separate)'' of Tab.~\ref{Tab:comparison_with_polyvit},
our model still shows a similar performance
even when separate (hence different) embedding layers for images and videos were used.
Results of our model with either image datasets only or video datasets only are 
also shown in Tab.~\ref{Tab:comparison_with_polyvit} in rows ``IV-ViT (images)'' and ``IV-ViT (videos)''.
The performance of IV-ViT for videos becomes better when training with videos only,
while the performance for images doesn't change when training with images only.
This suggests the need for more investigation on training videos when jointly trained with images.

Another comparison with PolyViT
shown in Table~\ref{Tab:comparison_with_polyvit} is the training from scratch.
For the fixed number of iterations, our IV-ViT is faster and easier to train.
As seen in Fig.~\ref{fig:PolyViT scratch}, the performance of PolyViT for UCF101 looks saturated indicating overfitting.
In contrast, Fig.~\ref{fig:ours scratch} shows that the performance increases steadily for all datasets,
which demonstrates the effectiveness of the architecture of IV-ViT.

\section{Conclusion}

In this paper, we propose IV-ViT, a model that can handle images and videos in the same model,
by adopting a frame-wise ViT with temporal feature shift.
Our IV-ViT is limited to image and video modalities, however, allowing extensions in several ways.
Future work includes the use of other temporal modeling such as
TimeSformer \cite{Bertasius_ICML2021_TimeSformer} or VideoSwin \cite{Liu_2022CVPR_VideoSwin}
instead of feature shift,
more sophisticated temporal fusion \cite{Neimark_2021_ICCVW_VideoTransformerNetwork,Sharir_arxiv2021_STAM} instead of average pooling,
and tuning training procedures and hyperparameters for each dataset.

\section*{Acknowledgement}
This work was supported in part by JSPS KAKENHI Grant Number JP22K12090.

\clearpage
\bibliographystyle{ieee_fullname}
\bibliography{mybib, all}

\end{document}